%% file: paper.tex
\documentclass[runningheads]{llncs}
\usepackage[T1]{fontenc}
\usepackage{graphicx}
\usepackage{booktabs}
\usepackage[misc]{ifsym}

\usepackage{mwe}

\input{packages}

\input{macros}

\definecolor{afiablue}{RGB}{61,159,207}
\definecolor{afiared}{RGB}{167,75,68}
\definecolor{afialightblue}{RGB}{158,193,232}

\definecolor{carnelian}{rgb}{0.7, 0.11, 0.11}

\definecolor{electricpurple}{rgb}{0.75, 0.0, 1.0}

\begin{document}
    
    \title{Physical knowledge on historical data matters more than enforcing physical constraints on the forecast}
    
    \titlerunning{Physics informed backcasting and forecasting}
    
    \author{
            Etienne Lehembre\inst{1}\orcidID{0000-0002-1374-5453} \and
            Pascal Audigane\inst{2}\orcidID{0000-0003-1445-8314} \and
            Vincent Nguyen\inst{1}\orcidID{} \and
            Christel Vrain\inst{1}\orcidID{0000-0003-3307-0753}\and
            Thi-Bich-Hanh Dao\inst{1}\orcidID{0000-0002-2740-6954} 
        }
    
    \authorrunning{E. Lehembre et al.}
    
    \institute{ 
        Université d’Orléans, INSA CVL, LIFO, UR 4022, Orléans, France \and
        BRGM, 3 avenue Calude Guillemin, 45060, Orléansn, France
    }
    
    \maketitle              

    \begin{abstract}
        Time series forecasting has seen significant advancements with the emergence of new deep learning models. However, forecasting time series in applications involving physical processes remains a major challenge. Despite the apparition of Physics Informed Neural Networks (PINN), recent models do not estimate unobservable intermediate physical variables, which are important for domain experts to understand the target behavior. To this end, we propose a Physics Informed Recurrent Neural Network (PIRNN) which predicts, along the target, unobservable variables on both historic data and forecast target. This approach enhances the model robustness and results interpretation using domain knowledge. Our method is easily adaptable to any physical model using several equations, each having its own set of unobservable variables, to describe it-self. As a case study, we incorporate physical equations used for groundwater levels predictions by the physical model called Gardenia. This model uses transfers equations between reservoirs, optimized with data assimilation, to simulate the evolution of groundwater levels. Evaluation includes several well known neural network models and the Gardenia model compared on twelve real world datasets. In addition, we study the impact of each component through an ablation study. Our model outperforms other models on five out of the twelve datasets and our ablation study underlines the importance of having a physical background in our time series forecasting task. Finally, the coherence of the physical variables predicted by our neural network is assessed by a domain expert.
    
    \end{abstract}

    \input{00Introduction}

    \input{01RelatedWork}
    \input{02Preliminaries}
    \input{03Method}
    \input{04Experiences}

    \input{05Conclusion}
    
	\bibliographystyle{splncs04}
	\bibliography{bibliography}
	
\end{document}

%% file: packages.tex
\usepackage{todonotes}
\usepackage{mathtools}
\usepackage{amsmath}
\usepackage{amssymb}

\PassOptionsToPackage{hyphens}{url}
\usepackage{xurl}

\usepackage{algorithm}
\usepackage{algpseudocode}

%% file: macros.tex
\def\modelname{BFcastPIRNN}
\def\modelfullname{Backcast Forecast Physics Informed RNN}

\def\learningrate{0.001}

\def\gheight{G1}

\def\hheight{H}

\def\uheight{U}
\def\soillevel{\uheight}

\def\time{t}

\def\gsiphon{TG}
\def\hsiphon{TH}
\def\ruiper{RS}
\def\umax{\ensuremath{C_{g}}}

\def\soilcapacity{\umax}

\def\rain{\ensuremath{R}}
\def\truerain{\rain'}
\def\evapo{\ensuremath{E}}
\def\trueevapo{\evapo'}

\def\effectiverain{\ensuremath{{\rain_{E}}}}

\def\gderivative{\ensuremath{\frac{\partial \gheight}{\partial \time}}}
\def\hderivative{\ensuremath{\frac{\partial \hheight}{\partial \time}}}
\def\uderivative{\ensuremath{\frac{\partial \uheight}{\partial \time}}}

\def\gsiphonmax{\ensuremath{\gsiphon_{max}}}
\def\gsiphonmin{\ensuremath{\gsiphon_{min}}}
\def\hsiphonmax{\ensuremath{\hsiphon_{max}}}
\def\hsiphonmin{\ensuremath{\hsiphon_{min}}}

\def\timeserie{S}
\def\var{x}
\def\target{y}

\def\phyvar{\ensuremath{\mathcal{U}}}
\def\cstvar{\ensuremath{\mathcal{C}}}
\def\phyfun{\ensuremath{\delta}}

\def\phyderivatives{\ensuremath{\partial_{\phyvar}}}

\def\loss{\mathcal{L}}

\def\dataloss{\ensuremath{\loss_{Data}}}
\def\physicalloss{\ensuremath{\loss_{Physical}}}
\def\borderloss{\ensuremath{\loss_{Border}}}

\def\lossgsiphonmax{\ensuremath{\loss_{\gsiphonmax}}}
\def\lossgsiphonmin{\ensuremath{\loss_{\gsiphonmin}}}
\def\losshsiphonmax{\ensuremath{\loss_{\hsiphonmax}}}
\def\losshsiphonmin{\ensuremath{\loss_{\hsiphonmin}}}

\newcommand{\meanof}[1]{\mu({#1})}
\newcommand{\abs}[1]{||{#1}||}

\def\predictionsize{p}
\def\windowssize{w}

\def\historic{\ensuremath{x_{t-\windowssize:t}}}
\def\groundtruth{\ensuremath{\target_{t:t+\predictionsize}}}
\def\prediction{\ensuremath{\hat{\target}_{t:t+\predictionsize}}}

\def\backcast{\ensuremath{\hat{\phyvar}_{t-\windowssize:t}}}
\def\forecast{\ensuremath{\hat{\phyvar}_{t:t+\predictionsize}}}

\def\backcastfun{\ensuremath{f_b}}
\def\forecastfun{\ensuremath{f_f}}

\def\phypred{\ensuremath{\hat{\phyvar}_{t-\windowssize:t+\predictionsize}}}

\def\cstpred{\ensuremath{\hat{\cstvar}}}

\def\phyhistoric{\ensuremath{\phi_{x,{t-\windowssize:t}}}}

%% file: 00Introduction.tex
\section{Introduction}
\label{sec:intorduction}

Time series have long been studied~\cite{gershenfeld2018future} for analyzing processes that evolve over time in a wide range of domains, including medicine~\cite{piccialli2021artificial}, banking~\cite{winistorfer2023measure}, and resource management~\cite{oreshkin_n-beats_2021}.
With the increasing availability of computational power, the rise of neural networks brought powerful data-driven tools to time series forecasting~\cite{miller2024survey,li2024deep}. Yet, classical models typically offer limited explainability and often fail to incorporate domain knowledge such as physical equations. 
This lack of transparency stems from their complex architectures and the absence of explicit links between internal representations and physical or interpretable concepts.
Moreover, few statistical or purely neural approaches are designed to handle scenarios where key physical parameters are unknown or where systems are driven by irregular, uncertain exogenous variables.
In recent years, new models have emerged that incorporate physical equations into their training process~\cite{raissi_physics-informed_2019}, offering new possibilities for problems governed by physical laws.
This is a complex task, characterized by {1)} nonlinear dynamics governed by differential equations, and {2)} partially observable external variables not available on the forecasting period such as rainfall and temperature. Crucially, while the governing physical model (e.g., Ordinary Differential Equations or {ODE}) is known, some essential physical parameters remain unknown and must be learned.

In this work, we propose a new architecture for long-range forecasting of time-dependent physical systems and apply it to groundwater levels.
We integrate equations from a well-established groundwater model into a recurrent architecture that jointly forecasts the target and infers unobservable aquifer variables. This provides both accurate forecasts and interpretable physical signals, validated by a domain expert in Section~\ref{sec:experiences}.

As water resource management~\cite{aslam2018groundwater} becomes more challenging under climate change, we target a model that is accurate, reliable, and lightweight enough for deployment in a digital twin at each piezometer update.
To this end, we propose a hybrid framework that combines physical knowledge with recurrent neural networks, preserves the underlying ODE structure, and enables data-driven inference of unknown physical parameters.

Section~\ref{sec:related_work} reviews related forecasting and physics-informed methods.
Section~\ref{sec:preliminaries} introduces the notations.
Section~\ref{sec:method} presents our architecture, and Section~\ref{sec:physic_intro} details its groundwater instantiation.
Section~\ref{sec:experiences} reports datasets, settings, and results, and Section~\ref{sec:conclusion} concludes.

%% file: 01RelatedWork.tex
\section{Related work}
\label{sec:related_work}


\paragraph{\textbf{Time series prediction.}}
Due to their inherent ability to extract complex patterns directly from data, neural networks are now central in time series forecasting~\cite{miller2024survey,li2024deep}. NHITS~\cite{challu_nhits_2023} is notable for combining forecasting and backcasting, with interpretable trend, seasonality and noise decomposition and fast training. Transformers are also widely used; TimeXer~\cite{wang2024timexer} is a recent architecture handling exogenous variables. LLM-based forecasting has also emerged~\cite{tan2024language,abdullahi2025time}, firstly leveraging natural language through prompts~\cite{xue2022leveraging}. But the large parameter count and data requirements conflict with our objective of lightweight per-piezometer models for a digital twin.

\paragraph{\textbf{Physics informed neural network (PINN).}}
Recently, an increasing number of studies have explored the integration of physical models into deep learning architectures to address the challenges posed by limited training data and the need to be grounded in physical equations and to account for physical dependencies.
The use of equations was first introduced in neural ODEs~\cite{chen2018neural}.
Physical partially differentiable equations was introduced by~\cite{raissi_physics-informed_2019}, which proposed the use of physics-informed loss functions. Currently, ODEs are leveraged in two distinct ways:
1) they are used to generate synthetic data points by sampling random coordinates within the problem domain to train a neural network,
2) the loss function is augmented with a physics-based term quantifying the discrepancy between the predicted outputs and the expected physical behavior.
Both methods aim to compensate for the lack of training data by incorporating domain knowledge~\cite{farea2024understanding}.
Many extensions target heat diffusion, fluid dynamics and chemical data~\cite{zhang_physics-informed_2020,lu2021deepxde,ren_phycrnet_2022,cho_lstm-pinn_2022,nagda_pits_2024,cheng2021deep,Depina02012022,schiano_di_cola_investigating_2024}. 
%
%
Nevertheless, these approaches do not explicitly address time series modeling through neural networks incorporating physical equations. 
Time is used as an input for multi-layer perceptrons or convolutional neural networks, computing the learned physical equation outputs for a given set of parameters, rather than aiming to forecast multiple points values in time.
Moreover, none of these approaches deals with intermediate unobservable parameters, the used physical equations rely solely on the inputs and user defined fixed parameters. 
In our work, we developed a novel model aiming to integrate both the prediction of unobservable physical variables and the forecasting of multiple steps in time.

%% file: 02Preliminaries.tex
\section{Preliminary notions}
\label{sec:preliminaries}


\paragraph{\textbf{Time series.}}
A multivariate time series is a time series in which each \emph{time step} is associated with multiple values, each corresponding to a distinct variable.
We denote by $\timeserie$ be a \emph{multivariate time series} of length $n$ with $k$ variables, defined as $\timeserie = [\var_1, \var_2, \cdots, \var_{n-1}, \var_n]$, where each $\var_i = (\var^1_i, \var^2_i, \cdots, \var^k_i)$ represents the $k$-dimensional observation at time step $i$.



\paragraph{\textbf{Multi-Layer Perceptron (MLP) and Recurrent Neural Network (RNN).}} A multi-layer perceptron~\cite{rosenblatt_perceptron_1958} is a function defined as $f:\mathbb{R}^n \rightarrow \mathbb{R}^m$, where $n$ is the dimension of the input and $m$ the dimension of the output. A Long Short-Term Memory (LSTM) network~\cite{hochreiter_long_1997} is a variant of recurrent neural networks designed to capture long-term dependencies. At each time step $t$, the state of a LSTM cell is defined by a hidden state vector $h_t$ and a cell state vector $c_t$. A \emph{Gated Recurrent Unit} (GRU) network~\cite{cho2014propertiesgru} is a variant of the LSTM with only the hidden state vector $h_t$.

%% file: 03Method.tex
\section{\modelfullname}
\label{sec:method}


In this section, we present {\modelfullname} ({\modelname}), which integrates a physical model into a recurrent neural network for forecasting physics-guided time series.
The objective is to predict the target $\groundtruth$ from historical observations while estimating unobservable variables $\phyvar$ and constants $\cstvar$ involved in the physical function $\phyfun(\timeserie, \phyvar, \cstvar)=\target$.
Given a historical window $\historic$, the model forecasts $\predictionsize$ future points $\prediction$.

\subsection{Architecture insight}
\paragraph{}To model temporal dependencies with low computational cost, the processor can be either an LSTM or a GRU.
Both backcast and forecast blocks follow an encoder-processor-decoder design: MLP encoder, RNN processor, and MLP decoder.
The backcast block estimates latent variables on historical data, and the forecast block predicts future target values and latent variables.

\begin{figure}[ht]
    \centering
    \includegraphics[width=1.\linewidth]{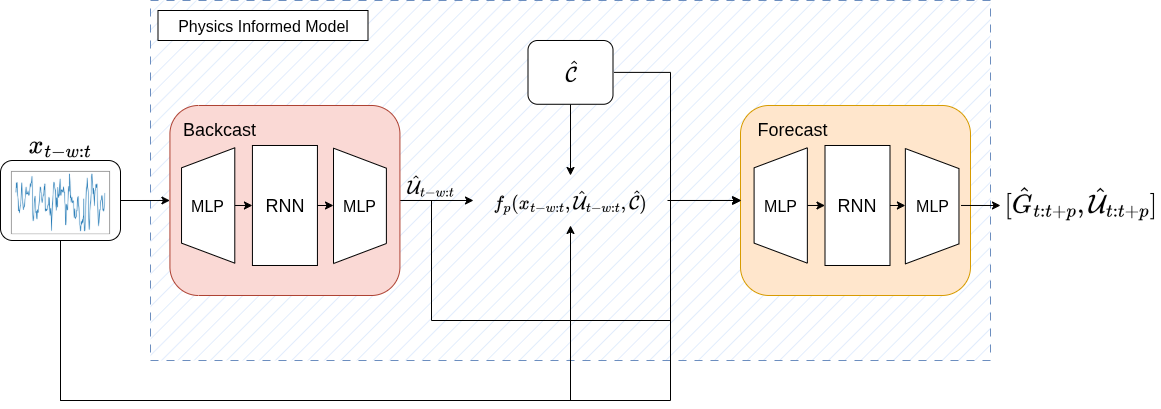}
    \caption{Summary of the {\modelfullname}. The backcast block is used to predict the unobservable physical variables needed to compute the physical covariables of historical data. The forecast block is used to predict both the target and the unobservable variables needed to compute the physical covariables of the forecasted time series.}
    \label{fig:pilstm}
\end{figure}

Figure~\ref{fig:pilstm} summarizes the architecture of our hybrid model. 1) Unknown physical constants $\cstpred$ are represented as learnable parameters.
2) A \emph{backcast block} reconstructs the unobservable variables $\backcast$ from the historical data $\historic$. 3) These reconstructed variables are then processed and combined with the historical data. 4) The resulting physics-informed representation is passed to the \emph{forecast block}, which predicts the target sequence $\prediction$  and the associated unobservable physical variables $\forecast$.
Both the backcast and forecast blocks follow an encoder - processor - decoder structure, using a RNN to process the temporal information in the latent space. According to Hamza{\c{c}}ebi et al.~\cite{hamzaccebi2009comparison}, all time steps are predicted using a single MLP layer.

\subsection{Formal description}
\paragraph{}Let $t$ a given time and $\windowssize$ the historical window size. The auto-regressive model is defined as: $\prediction = f(\historic, \phyhistoric)$ with $\historic$ the historical data and $\phyhistoric$ the physics guided historical data, as defined below.

\paragraph{\textbf{Backcast block.}} The backcast block $\backcastfun()$ in charge of predicting unobservable physical features for the forecast block is composed of three main components. The MLP encoder $e$, the MLP decoder $d$ and the LSTM $g$ used as interface between the encoder and the decoder. Therefore,  $\backcast = \backcastfun(\historic) = d(g(e(\historic)))$ with  $\backcast$ the estimated unobserved variables.

\paragraph{\textbf{Physical processing.}} Given $\historic$, and estimated physical parameters $\backcast$ and $\cstpred$. We use physical equations $\phyfun$ to compute the physics informed historic $\phyhistoric = [\phyfun(\historic, \backcast, \cstpred), \phyderivatives, \backcast, \cstpred]$ with $\phyderivatives$ the physical derivatives of $\phypred$ with regard of $\phyfun$. This physics informed historic is used to influence the prediction of the forecast block by untangling the underlying physical variables of the time series.
In this model, constants $\cstvar$ are model learnable parameters and $\phyvar$ are unobservable variables.

\paragraph{\textbf{Forecast block.}} The forecast block $\forecastfun()$ predicting the future steps of the time series is composed of three main components. The MLP encoder $e$, the MLP decoder $d$ and the LSTM $g$. 
It takes as input $z_{t-\windowssize:t} = [\historic, \phyhistoric]$ the physics augmented historical data with the unobservable variables and the physical derivatives.
Therefore,  $\prediction = \forecastfun(z_{t-\windowssize:t}) = d(g(e(z_{t-\windowssize:t})))$ with  $\prediction$ the target forecast prediction and $\phypred$ the unobservable physical covariables allowing to study the forecast learned physical dynamic.

\paragraph{\textbf{Loss function}}
The loss function is defined by the mean squared error:
\begin{equation}
    \dataloss = \frac{1}{\predictionsize}\underset{i=0}{\overset{\predictionsize}{\sum}}(\hat{y}_{t+i} - y_{t+i})^2
    \label{eq:dataloss}
\end{equation}
Along with the residual error, {\modelname} incorporates a physical loss that enforces physical relations and a border loss that constrains domain boundaries to prevent deviant predictions.
One main challenge is to avoid the vanishing of any of the losses while preventing the total loss $\loss$ explosion.
Therefore, the loss of our physics informed model is expressed as follows:
\begin{equation}
    \loss = \alpha\dataloss + \beta\physicalloss + \gamma\borderloss
    \label{eq:loss}
\end{equation}
Where $\alpha, \beta$ and $\gamma$ are either predefined user parameters or dynamically optimized parameters adjusting the balance between the loss function~\cite{bischof2025multi}.

\paragraph{\textbf{{\modelname} variants.}} Our model comes in three variations. The full model we described above called {\modelname}, a model without a physics informed backcast called FcastPIRNN and a model without physics features for the forecast block called BcastPIRNN. In the FcastPIRNN, the backcast creates a latent space focused on the time dependencies in the historical data on which, no loss is applied. Nevertheless, the forecast block is still trained with a mixture of data loss, physical loss and border loss. Regarding the BcastPIRNN, only the backcast block is trained based on the physical features. To smoother the convergence, the forecast block does not predict any physical feature, only the target. While the prediction is less constrained by the physical properties, it still takes into account the physical dynamics passed by the backcast block. In the BcastPIRNN case, the loss function is a combination of physics informed loss on the backcast and data loss on the forecast. In all of these three models, unobservable physical constants and variables are still learned to model the given time series. Our model, and its variants, can be applied for a time series forecast problem under physical laws, in order to learn unobservable parameters. In Section \ref{sec:physic_intro} we present its application to groundwater forecasting.

\section{Physic guided groundwater forecasting}
\label{sec:physic_intro}
Groundwater level prediction is crucial for water management but remains a complex task as it is influenced by meteorological events, soil composition and fluid mechanism.

\paragraph{\textbf{Physical model for groundwater prediction.}}
Accurate and reliable groundwater level forecasting remains vital to correctly address water resource management in several contexts (impact of climate change, sustainability of drinkable water, industrial uses, ...). Among simulation tools, we consider \emph{Gardenia}~\cite{thiery_gardenia_2014} a physical model developed by the French geological survey \emph{BRGM}\footnote{\url{https://www.brgm.fr/fr}} and recently deployed as 
Rameau~\footnote{\url{https://rameau.readthedocs.io/en/latest/index.html}}.
Gardenia is a physical model modeling timeseries of groundwater level variations using a set of reservoirs, which reproduces the path of water coming from precipitation to percolation through the ground, to reach an aquifer. Gardenia uses a modified version of the Rosenbrock method~\cite{rosenbrock1960automatic} to optimise physical parameter values by confronting simulated hydrological variables with observations, i.e. using data assimilation to solve an inverse problem on the desired ODEs.
    \begin{figure}[ht]
        \includegraphics[width=0.8\linewidth]{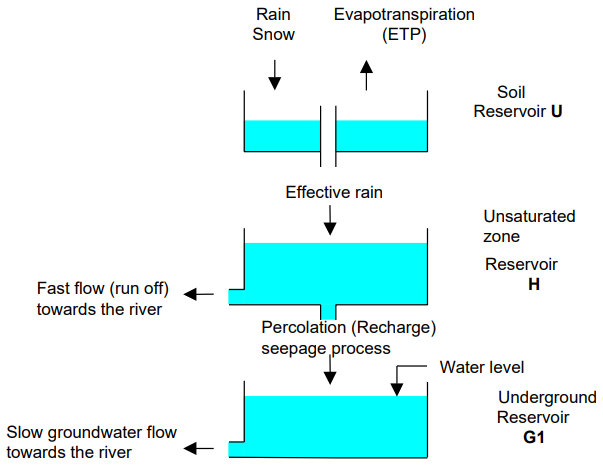}
        \centering
        \caption{Groundwater model with one tank.}
        \label{fig:gardenia_1}
    \end{figure}
    Figure~\ref{fig:gardenia_1} shows the decomposition of a hydrosystem divided into three reservoirs: $\uheight$ for the soil, $\hheight$ for the unsaturated zone, and $\gheight$ for the fully saturated zone. 
    The first reservoir $\uheight$ controls the hydrometric balance between the rain falling on the ground, which is partly absorbed by the evapotranspiration (ETP) process.   
    The remaining rain reaching the unsaturated zone is called effective rain and can be calculated with the GR3 or the Thornthwaite model~\cite{thiery_gardenia_2014}, which is controlled by the maximal capacity factor $\umax$ of the soil reservoir. 
    The second reservoir, $\hheight$, is a transfer reservoir between the effective rain and the groundwater reservoir. A repartition process is controlled by the $\ruiper$ factor (runoff versus seepage processes).
    The third reservoir, $\gheight$, is the reservoir corresponding to the groundwater level.
    Both reservoirs, $\hheight$ and $\gheight$, exhibit an outflow to model the movement of water through soils with a kinetic rate controlled by the reservoir half-life $\gsiphon$, and $\hsiphon$, defined as the time required for the reservoir to empty by half in the absence of rainfall.

In the following, $\uheight$, $\hheight$ and $\gheight$ will be used to describe their reservoir and their water level. Then, the water level is the target $\target = \gheight$, physical unobservable features are reservoirs' level and effective rain $\phyvar = \{\uheight, \effectiverain, \hheight\}$ and constant features are the reservoirs' half life, the transfer reservoir point of equilibrium and the maximum level of the soil reservoir $\cstvar = \{\umax, \hsiphon, \ruiper, \gsiphon\}$ which are not time dependent.

The water level recorded is controlled by Equation~(\ref{eq:g_tank}) corresponding to the variations of the reservoir $\gheight$.
\begin{equation}
    \gderivative = \gheight * exp(\frac{-\time}{\gsiphon / ln(2)})
    \label{eq:g_tank}
\end{equation}
Where $d\time$ is the time step, $\gsiphon$ is the siphon factor of the tank $\gheight$, $\frac{\partial \gheight}{\partial \time}$ is the evolution of the $\gheight$ reservoir and $\frac{\gheight * d\time}{\gsiphon}$ is the outflow of $\gheight$.

The water level also depends of the reservoir $\hheight$ which evolution follows Equation~(\ref{eq:h_tank}).
\begin{equation}
    \hderivative = \frac{C * \ruiper * exp(\frac{-\time}{\hsiphon / ln(2)})}{1 - C * exp(\frac{-\time}{\hsiphon / ln(2)})}
    \label{eq:h_tank}
\end{equation}
Where $\effectiverain$ is the effective rain, $\hsiphon$ is the siphon factor of the reservoir $\hheight$§. The run-off seepage factor $\ruiper$ is the water level of the reservoir $\hheight$ for which the outflow $\frac{\hheight^2 * d\time}{\hsiphon * \ruiper}$ is equal to the percolation $\frac{\hheight * d\time}{\hsiphon}$ giving the flow from $\hheight$ to $\gheight$. We use the temporary variable $C= \frac{1}{(1+\frac{\ruiper}{\hheight})}$.

The effective rain is computed by the soil reservoir which is ruled by the GR3J model \cite{beauce_edijatno_1999} with Equation~(\ref{eq:u_tank}).
\begin{equation}
    \uderivative = 
    \begin{cases}
        \frac{\soillevel + \soilcapacity\tanh!\left(\frac{\truerain}{\soilcapacity}\right)}{1 + \left(\frac{\soillevel}{\soilcapacity}\right)\tanh!\left(\frac{\truerain}{\soilcapacity}\right)} & \text{if } \rain \ge \evapo \\
        \frac{1-\tanh!\left(\frac{\trueevapo}{\soilcapacity}\right)}{1+\left(1-\frac{\soillevel}{\soilcapacity}\right)\tanh!\left(\frac{\trueevapo}{\soilcapacity}\right)} & \text{if } \rain < \evapo
    \end{cases}
    \label{eq:u_tank}
\end{equation}
Where $\umax$ is the capacity threshold of the $\uheight$ reservoir, $\truerain = \rain - \evapo$ and $\trueevapo = 0$ is the remaining rainfall and the $\trueevapo = \evapo - \rain$ the remaining evapotranspiration.

\paragraph{\textbf{Physics informed loss functions.}} To use the equations of Gardenia, the neural network has to predict every parameter that cannot be observed in data but depends of Equations~(\ref{eq:g_tank}) and (\ref{eq:h_tank}). 
Using the Gardenia physical model, Equations~(\ref{eq:g_tank}), (\ref{eq:h_tank}) and (\ref{eq:u_tank}) are transformed into a loss function~(\ref{eq:physicalloss}), which  acts as a soft constraint on the neural network, enforcing convergence towards a function compatible with the physical laws of groundwater reservoirs.
\begin{equation}
    \begin{aligned}
        \physicalloss = 
        & \frac{1}{\windowssize+\predictionsize}\underset{i=0}{\overset{\windowssize+\predictionsize}{\sum}}((\uderivative(i) - 
        \begin{cases}
        \frac{\soillevel(i) + \soilcapacity\tanh!\left(\frac{\truerain(i)}{\soilcapacity}\right)}{1 + \left(\frac{\soillevel(i)}{\soilcapacity}\right)\tanh!\left(\frac{\truerain(i)}{\soilcapacity}\right)} & \text{if } \rain(i) \ge \evapo(i) \\
        \frac{1-\tanh!\left(\frac{\trueevapo(i)}{\soilcapacity}\right)}{1+\left(1-\frac{\soillevel(i)}{\soilcapacity}\right)\tanh!\left(\frac{\trueevapo(i)}{\soilcapacity}\right)} & \text{if } \rain(i) < \evapo(i)
        \end{cases}\\
        & +\frac{1}{\windowssize+\predictionsize}\underset{i=0}{\overset{\windowssize+\predictionsize}{\sum}}(\hderivative(i) - \frac{C(i) * \ruiper * exp(\frac{-\time}{\hsiphon})}{1 - C(i) * exp(\frac{-\time}{\hsiphon})})^2 \\
        & + \frac{1}{\windowssize+\predictionsize}\underset{i=0}{\overset{\windowssize+\predictionsize}{\sum}}(\gderivative(i) - \gheight(i) * exp(\frac{-\time}{\gsiphon}))^2 
    \end{aligned}
\label{eq:physicalloss}
\end{equation}
Where $\windowssize$ is the historic size, $\predictionsize$ is the forecast horizon, $\gheight(i)$ for $i\in[0, \windowssize[$ is historic groundwater levels, $\gheight(i)$ with $i\in[\windowssize, \windowssize+\predictionsize[$, $\hheight(i)$ and $\soillevel(i)$ are predicted reservoir levels, and $\gsiphon$, $\hsiphon$,, $\ruiper$ and $\soilcapacity$ are the constant physical factors.

In order to have consistent outputs, we add a border loss which goal is to set a weak constraint on the minimum value and the maximum value of $\gheight$ and $\hheight$ {(as in the Gardenia model)} with hyper-parameters giving four losses: $\lossgsiphonmax, \lossgsiphonmin, \losshsiphonmax$ and $\losshsiphonmin$.
To give an example of these functions we detail $\lossgsiphonmax = \frac{1}{N}\underset{i=0}{\overset{\predictionsize}{\sum}}min(0, \gsiphonmax - \gsiphon_i)^2$. 

With the above loss functions, we are now able to define our border loss with Equation~(\ref{eq:borderloss}):

\begin{equation}
    \begin{aligned}
        \borderloss = & \lossgsiphonmax + \lossgsiphonmin + \losshsiphonmax + \losshsiphonmin
    \end{aligned}
    \label{eq:borderloss}
\end{equation}

By leveraging previously defined equations, our model introduced in Section~\ref{sec:method} can be applied to forecast groundwater levels within the constraints of the equations of the Gardenia model.


%% file: 04Experiences.tex
\section{{Experiments} and results}
\label{sec:experiences}
    In this section,  we compare our model to several machine learning models, all trained on monthly data. We address three questions: 
    \begin{enumerate}
        \item Does {\modelname} outperform baselines ?
        \item Which configuration performs best ?
        \item Are inferred unobservable variables consistent with expert knowledge ?
    \end{enumerate}
    
    \subsection{Materials and Methods}
    Datasets merge groundwater levels (\url{https://ades.eaufrance.fr/}), meteorological variables (\url{https://www.data.gouv.fr/datasets/donnees-climatologiques-de-base-mensuelles}), and evapotranspiration (\url{https://www.data.gouv.fr/datasets/etp-fao-hargreaves}).
    The ETL pipeline is available at \url{https://github.com/TronnoxUwU/STAGE-Junon}, and model code with hyper-parameters at \url{https://github.com/Etienne-Lehembre/BackcastForecastPIRNN}.
    
    \paragraph{\textbf{Studied area and data management.}}
    We focus our study on one watershed connected to the Loire river. The aquifer water level is monitored by several tens of piezometers~\cite{beauce_serviere_2025}. 
   In order to have datasets with enough daily measurement we concentrate our period of study between 1995 and 2018. 
    We observed in our data some cyclicity in the variation which is related to two main physical phenomena. The first is an annual seasonal cycle representing the variation of rain and vegetation along the year;
    this is observed for almost all wells. 
    The second phenomena is related to the heterogeneity of the underground and especially the unsaturated zone. 
    The capacity of the underground to store water or to exchange laterally with other aquifers nearby can induce a slower variation of the water level with a longer period (about 12 years for the studied area). 
    Such hydrosystems are called \emph{inertial} meaning that they react hydrodynamically with a slower process than \emph{reactive} aquifers which are evolving annually.
    
    In order to evaluate the impact of the use of physical equations on learning, we select $12$ piezometers with the \emph{water level} as target and the \emph{rain} and the \emph{evapotranspiration} as exogenous variables.
    In order to better observe the ability of models to learn from several datasets, we split the $12$ data sets into two groups.
    \begin{itemize}
        \item The inertial group witth a cyclicity of at least 12 years: 02923X0007\_F (282 - 12), 03266X0009\_P (278 - 12), 03622X0027\_PZ (288 - 12), 02936X1018\_P (278 - 12), 02936X2005\_PFAEP (288 - 12), 03272X0006\_PZ (288 - 12), 03614X0001\_PAEP (288 - 12) and 03631X0099\_F (287 - 12).
        \item The reactional group with a cyclicity of 1 year: 03288X0042\_P (286 - 1), 03983X0267\_PZ3 (288 - 1), 04312X0039\_F (287 - 1) and 04293X0003\_FAEP (283 - 1).
    \end{itemize}
    Therefore, the models trained on multiple datasets are trained on datasets having similar cyclicity and physical dynamics.
    
    \paragraph{\textbf{Data preprocessing.} }
    Each time series is separated into train, evaluation and test. The last 5 years are divided between evaluation (4 years) and test (last year).
    Data gaps are filled with a first degree interpolation, then datasets are aggregated with a monthly granularity by mean. After aggregation, datasets are normalized between $0$ and $1$ with a min max normalization fitting the training part of the time series.
    Each data set is transformed in a list of couples $(\historic, \groundtruth)$ with $\historic$ the historical data and $\groundtruth$ the future ground truth.
    The length of the historical data varies between 12, 36 and 60 months, while the forecasting period is set to 12 months.
    Each model is trained on the shuffled list of couples composed either of all datasets for a given group or of one dataset. For each couple, we get the model forecasting and compare it to the groundtruth $y_{\time:\time+\predictionsize}$. 

    \paragraph{\textbf{Baseline models.}}
    We choose N-BEATS~\cite{oreshkin_n-beats_2021}, NHITS~\cite{challu_nhits_2023} and TimeXer~\cite{wang2024timexer} run with pytorch forecasting\footnote{\url{https://pytorch-forecasting.readthedocs.io/en/stable/}} as state-of-the-art models for they are both lightweight and easily trainable.
    We also compare {\modelname} to a LSTM and a GRU network based on the forecast block architecture and to Gardenia which is our physical guideline model.
    
    \paragraph{\textbf{Gardenia simulation process.}} The expert performed a set of simulation of the water level variation on daily data between 1995 and 2018 using the following methodology: 1) first the water level of the three reservoirs is initiated  during a cycle of 8 years using a fictive rain corresponding to the average value of the rain in the time series (1995-2018), then 2) the parameters  which control the simulation process (progressive capacity of the reservoir $\uheight$, siphon factors $\gsiphon$, $\hsiphon$, and the $\ruiper$ factor that defines the ratio between seepage and run-off using the data between 1996 and 2013) {are optimized}, 3) finally the five last year{s} of data {are simulated} with the optimised parameters, the rain and the evapotranspiration. Here, step 2) correspond to the training dataset and step 3) to evaluation with the test year included in it. All metrics are computed on monthly aggregation of both data and model prediction.
    
    \paragraph{\textbf{Physical models.}} Our model is declined in several versions. The RNN tested are: GRU, bidirectional GRU (bi-GRU), LSTM and bidirectional LSTM (bi-LSTM). The physical constraints combinations are: backcast only (BcastPIRNN), forecast only (FcastPIRNN) and both (BFcastPIRNN).
    
    \paragraph{\textbf{Evaluation metrics.}} 
    In order to evaluate our models, we rely on metrics presented by~\cite{dawson2007hydrotest} in their analysis of evaluation techniques for hydrological forecasting.
    Mean Absolute Error (MAE), $MAE(y_i, \hat{y_i}) = \frac{1}{\predictionsize}\underset{i=0}{\overset{\predictionsize}{\sum}}\abs{y_i - \hat{y}_i}$, measures average deviations.
    Root Mean Squared Error (RMSE), $RMSE(y_i, \hat{y_i}) = \sqrt{\frac{1}{\predictionsize}\underset{i=0}{\overset{\predictionsize}{\sum}}(y_i - \hat{y}_i)^2}$, emphasizes larger errors.
    Nash-Sutcliffe efficiency (NSE), $NSE(y_i, \hat{y_i}) =  1- \frac{\underset{i=0}{\overset{\predictionsize}{\sum}}(y_i - \hat{y}_i)^2}{\underset{i=0}{\overset{\predictionsize}{\sum}}(y_i - \meanof{y})^2}$, evaluates predictive skill against a mean baseline.
    MAE and RMSE are minimized, whereas NSE is maximized.

    \paragraph{\textbf{Optimizer and hyper-parameters.}} 
    Neural-network experiments are run with PyTorch Forecasting.
    We use RADAM~\cite{liu2019variance} with learning rate $\learningrate$, seed 42, batch size 32, maximum 250 epochs, and patience 10.
    All processors use two RNN layers.
    
    \subsection{Result analysis}
    To better understand results, it is important to remember that the experimentation methodology of Gardenia only divide datasets between train and evaluation. Therefore, we compare here {the results of our} machine learning models {obtained on the test dataset} with Gardenia results {on its evaluation dataset}, which are the same datasets. 

    \begin{table*}[ht]
        \centering
        \setlength{\tabcolsep}{3pt}
        \caption{Best forecasting performance (NSE/MAE/RMSE) for each model (rows) and datasets (columns).}
        \label{tab:perf_comp_metrics}
        \resizebox{\columnwidth}{!}{
            \begin{tabular}{lcccc}
                \hline
                Model & 02923X0007\_F & 03266X0009\_P & 03622X0027\_PZ & 02936X1018\_P \\
                \hline
                Gardenia & -90.057/2.836/2.839 & -9.032/0.914/0.930 & \textbf{0.806}/\textbf{0.174}/\textbf{0.207} & -40.852/1.268/1.291 \\
                PIRNN (mono) & \textbf{0.861}/\textbf{0.099}/\textbf{0.109} & -0.131/0.262/0.314 & -2.048/0.707/0.830 & -3.302/0.342/0.379 \\
                PIRNN (multi) & -1.057/0.367/0.421 & -0.034/0.239/0.300 & 0.030/0.389/0.468 & \textbf{0.144}/\underline{0.130}/\textbf{0.169} \\
                N-BEATS (mono) & \underline{-0.552}/0.337/\underline{0.371} & -0.151/0.242/0.315 & -0.077/0.381/0.489 & -1.462/0.245/0.313 \\
                N-BEATS (multi) & -6.401/0.766/0.809 & -0.030/0.234/0.298 & 0.070/0.404/0.454 & -2.034/0.288/0.348 \\
                NHITS (mono) & -2.187/0.437/0.531 & -0.193/0.242/0.321 & -1.826/0.643/0.792 & 0.066/\textbf{0.124}/0.193 \\
                NHITS (multi) & -3.028/0.515/0.597 & -0.183/0.241/0.319 & -1.433/0.581/0.735 & \underline{0.078}/\textbf{0.124}/\underline{0.192} \\
                TimeXer (mono) & -7.089/0.807/0.846 & -0.402/0.298/0.348 & -0.078/0.414/0.489 & -1.221/0.249/0.297 \\
                TimeXer (multi) & -2.340/0.490/0.544 & \textbf{0.239}/\textbf{0.214}/\textbf{0.256} & \underline{0.302}/\underline{0.292}/\underline{0.394} & -0.799/0.230/0.268 \\
                RNN (mono) & -68.273/2.400/2.441 & -0.725/0.333/0.388 & -2.123/0.713/0.840 & -96.430/1.777/1.802 \\
                RNN (multi) & -0.679/\underline{0.257}/0.380 & \underline{0.089}/\underline{0.231}/\underline{0.282} & 0.022/0.417/0.470 & -2.012/0.305/0.317 \\
                \hline
                \hline
                Model & 02936X2005\_PFAEP & 03272X0006\_PZ & 03614X0001\_PAEP & 03631X0099\_F \\
                \hline
                Gardenia & -526.744/2.261/2.289 & -660.349/2.387/2.392 & -14.209/0.675/0.719 & -42.529/1.710/1.719 \\
                PIRNN (mono) & \textbf{0.301}/\textbf{0.079}/\textbf{0.090} & -5.247/0.177/0.215 & -0.001/0.170/0.208 & -2.570/0.442/0.499 \\
                PIRNN (multi) & -8.163/0.298/0.326 & -1.298/0.117/0.130 & \underline{0.262}/0.147/\underline{0.178} & -0.079/0.217/0.274 \\
                N-BEATS (multi) & -8.820/0.294/0.312 & -5.104/0.212/0.230 & -0.001/\underline{0.142}/0.185 & 0.126/0.222/0.244 \\
                N-BEATS (mono) & -7.194/0.266/0.285 & -3.796/0.183/0.204 & -0.138/0.144/0.197 & \underline{0.153}/\underline{0.214}/\underline{0.240} \\
                NHITS (mono) & -1.035/0.103/0.142 & \textbf{0.235}/\textbf{0.067}/\textbf{0.081} & -0.064/0.159/0.190 & -2.312/0.398/0.474 \\
                NHITS (multi) & \underline{-0.852}/\underline{0.102}/\underline{0.136} & \underline{0.157}/\underline{0.068}/\underline{0.085} & -0.048/0.156/0.189 & -2.301/0.396/0.473 \\
                TimeXer (mono) & -7.470/0.262/0.290 & -10.337/0.276/0.313 & \textbf{0.365}/\textbf{0.120}/\textbf{0.147} & -0.192/0.227/0.285 \\
                TimeXer (multi) & -7.725/0.274/0.294 & -0.599/0.105/0.118 & -0.317/0.183/0.212 & \textbf{0.570}/\textbf{0.144}/\textbf{0.171} \\
                RNN (mono) & -521.029/2.450/2.458 & -928.797/2.607/2.619 & -1.004/0.215/0.294 & -42.757/1.738/1.745 \\
                RNN (multi) & -76.567/0.916/0.947 & -12.949/0.292/0.321 & 0.076/0.154/0.199 & -0.134/0.217/0.281 \\
                \hline
                \hline
                Model & 03288X0042\_P & 03983X0267\_PZ3 & 04312X0039\_F & 04293X0003\_FAEP \\
                \hline
                Gardenia & \textbf{0.911}/\textbf{0.599}/\textbf{0.714} & -1.131/0.719/0.733 & 0.483/1.223/1.517 & -0.291/1.072/1.276 \\
                PIRNN (mono) & \underline{0.599}/\underline{0.960}/\underline{1.585} & \textbf{0.485}/\textbf{0.244}/\textbf{0.361} & \underline{0.767}/0.750/\underline{1.001} & \textbf{0.835}/\textbf{0.385}/\textbf{0.451} \\
                PIRNN (multi) & 0.128/1.942/2.339 & \underline{0.481}/\underline{0.296}/\underline{0.362} & 0.499/1.257/1.467 & 0.316/0.748/0.918 \\
                N-BEATS (mono) & 0.018/2.067/2.373 & -0.812/0.565/0.676 & 0.593/1.054/1.346 & 0.230/0.752/0.986 \\
                N-BEATS (multi) & -0.155/2.193/2.572 & -0.478/0.515/0.611 & -0.012/1.894/2.122 & -0.036/0.955/1.143 \\
                NHITS (mono) & -0.817/2.524/3.227 & -0.137/0.455/0.536 & -0.105/1.804/2.217 & -0.109/0.926/1.183 \\
                NHITS (multi) & -1.090/2.709/3.461 & -0.078/0.443/0.522 & -0.085/1.804/2.197 & -0.077/0.926/1.166 \\
                TimeXer (multi) & -0.031/1.895/2.431 & 0.173/0.400/0.457 & 0.728/0.857/1.101 & 0.588/0.562/0.721 \\
                TimeXer (mono) & -0.006/1.999/2.401 & -0.015/0.424/0.506 & \textbf{0.835}/\underline{0.723}/\textbf{0.858} & 0.616/0.533/0.696 \\
                RNN (mono) & 0.217/1.536/2.216 & -1.146/0.649/0.737 & 0.646/\textbf{0.693}/1.234 & \underline{0.774}/\underline{0.436}/\underline{0.529} \\
                RNN (multi) & 0.040/2.124/2.454 & \textbf{0.485}/0.313/\textbf{0.361} & 0.208/1.582/1.845 & 0.236/0.778/0.971 \\
                \hline
            \end{tabular}
        }
    \end{table*}

    \begin{table}
        \centering
        \caption{Percentage of forecasting improvement of the PIRNN regarding each baseline model.}
        \label{tab:improvement_percent}
        \begin{tabular}{ccccc}
            \hline
            Gardenia & N-BEATS & NHITS & TimeXer & RNN \\
            \hline
            83.33\% & 75.00\% & 91.67\% & 50.00\% & 91.67\% \\
            \hline
        \end{tabular}
        
    \end{table}

    \paragraph{\textbf{1. Forecasting improvement.}} Tables~\ref{tab:perf_comp_metrics} report NSE, MAE and RMSE for each model (rows) and each dataset (columns), with values given in the format NSE/MAE/RMSE in each cell. Models trained on one dataset have the mention \textit{(mono)} while models trained on several datasets have the mention \textit{(multi)}. 
    Concerning our model (PIRNN) and the baseline RNN, mono and multi rows report the best configuration per test dataset.
    The main result is that PIRNN is the most consistent top performer across datasets and metrics. In particular, PIRNN ranks first on 5 out the 12 datasets and second on 3 out the twelve datasets, making it the strongest model overall in this benchmark.
    Regarding the NSE, the PIRNN shows the most consistent performance across datasets, with 8 out of 12 datasets having a positive NSE.
    These results show that physics-informed constraints provide robust forecasting quality across both mono- and multi-dataset settings.
    Other methods can still dominate specific datasets: TimeXer (mono/multi) leads on three datasets, NHITS (mono) is best on 03272X0006\_PZ, and Gardenia remains best on 03622X0027\_PZ and 03288X0042\_P for the three metrics. However, taken globally, the PIRNN remains the reference model. 
    In particular, Table~\ref{tab:improvement_percent} shows that PIRNN improves forecasting performance by 83.33\% regarding Gardenia, by 75.00\% regarding N-BEATS, by 91.67\% regarding NHITS, by 50.00\% regarding TimeXer and by 91.67\% regarding RNN. Moreover, while improving the forecasting performance, PIRNN also provides physical interpretability of the results by estimating the unobservable physical variables and constants.

    \begin{table}[ht]
        \centering
        \caption{Comparison of forecasting performance between physical model types (rows) and RNN model types (columns) based on mean (min - max) RMSE. Best values are in \textbf{bold} and second best are \underline{underlined}.}
        \resizebox{\columnwidth}{!}{
            \begin{tabular}{rcccc}
                \hline
                Variant & GRU & bi-GRU & LSTM & bi-LSTM \\
                \hline
                BFcastPIRNN & \textbf{1,164} (0,288 - 2,565) & 1,409 (0,388 - 2,943) & \underline{1,021} (0,212 - 3,315)  & \underline{1,084} (0,312 - 2,751)\\
                BcastPIRNN & \underline{1,217} (0,367 - 2,513) & \textbf{0,792} (0,090 - 2,374)) & \textbf{0,845} (0,130 - 2,588) & \textbf{0,789} (0,215 - 2,505)\\
                FcastPIRNN & 1,655 (0,208 - 3,578) & 1,832 (0,306 - 3,766) & 1,551 (0,251 - 3,100) & 1, 222 (0,253 - 3,113)\\
                RNN & 1,230 (0,226 - 2,906) & \underline{1,219} (0,274 - 2,795) & 1,216 (0,199 - 3,036) & 1,504 (0,357 - 3,464)
            \end{tabular}
        }
        \label{tab:mean-mae-rmse}
    \end{table}
    
    \paragraph{\textbf{2. Best configuration.}} Table~\ref{tab:mean-mae-rmse} shows mean (min - max) RMSE values across all datasets, grouped by variant type (rows) and RNN architecture (columns).
    The GRU consistently delivers the weakest performance, whereas the bi-GRU and bi-LSTM provide better predictions and are close match with the BcastPIRNN variant. Applying constraints solely to the backcast blocks yields the best overall results, although models with backcast and forecast constraints wins come in close second. We infer that providing a physical background is essential to improve the forecasting performances. Moreover, focusing the physical constraints only on the forecasting values leads to reduced performances.
    
    \begin{figure}[ht]
        \centering
        \includegraphics[width=0.9\linewidth]{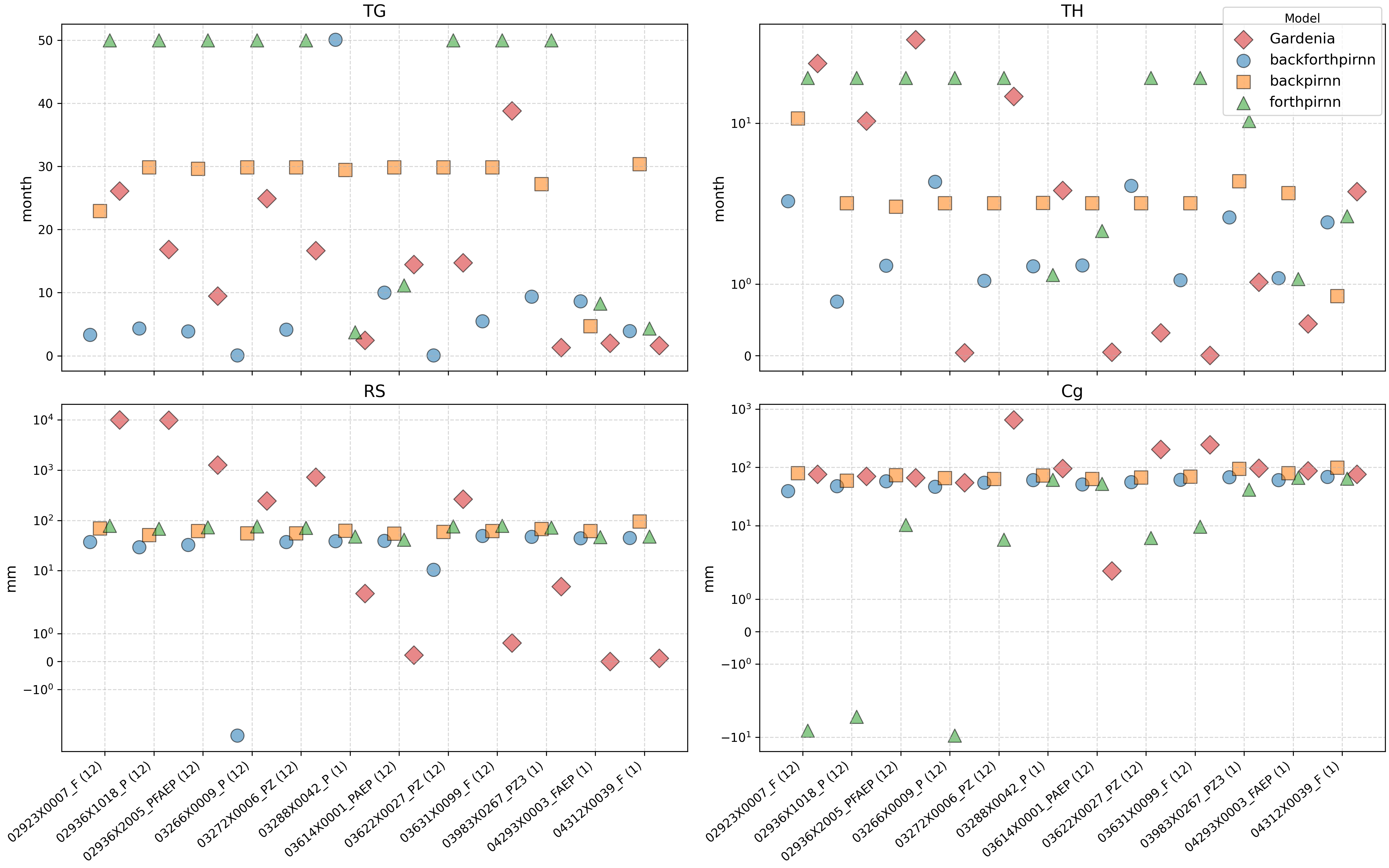}
        \caption{Water reservoir (x-axis) transfer reservoir half-life (y-axis) for each model (shape and color) and each data (label).}
        \label{fig:half-lives}
    \end{figure}
    
    \paragraph{\textbf{3. Physical interpretation.}} {The following results have been analyzed and validated by domain experts.} The range of values obtained for the estimations of the siphon factors  $\gsiphon$ and $\hsiphon$, the run-off seepage factor $\ruiper$ and the soil capacity factor $\soilcapacity$ are in good agreement with the Gardenia's optimisation results. As for Gardenia, we see in Figure~\ref{fig:half-lives} that BFcastPIRNN models tends to better differentiate the cyclicity of the time series with the $\gsiphon$ (lower for 1 year cycle, higher for the 12 years) and $\hsiphon$ factors. 
    On the other hand, the Back and the Forth models seems to provide the similar values for inertial time series, suggesting less reliable physical results for these models.
    Among the PIRNN models, we notice that the FcastPIRNN tends to overestimate the water reservoirs and transfer reservoirs of slow cyclicity datasets.
    Moreover, the predicted physical variables and constants allow us to reconstruct the water level time series using the water base level, rain and evapotranspiration, meaning that PIRNN successfully modeled the water reservoir system.
    Finally, PIRNN is also able to provide groundwater predictions without precipitations data (required for Gardenia simulations).
    
    

%% file: 05Conclusion.tex
\section{Conclusion}
\label{sec:conclusion}

We introduce in this paper {\modelfullname} (\modelname) and its variants, a generic machine learning model incorporating partially differentiable equations trained with data loss and physical loss to forecast one time series target and its correlated unobservable physical variables. Our model is designed to reconstruct and take advantage of unobservable physcial variables on the historical data and forecasted window to use them as an insight on the underlying physical dynamics of the time series.

We evaluate our model on several groundwater level time series with both inertial and responsive dynamics. {\modelname} reasonably outperforms both state-of-the-art machine learning methods and physical models regarding mean absolute error and root mean square error. 
In addition to the improvement of the forecasting results, our model outputs several physical variables allowing the experts to analyse the consistency of the model.
When examining the three variants of {\modelname}, the version that applies physics constraints only to the forecast appears to be the least effective in both forecasting performance and physical reliability. This indicates that restricting the physics-informed loss to the forecast alone may result in less reliable models.

Thanks to these lightweight physics informed model, we are able to include in a digital twin, fast trainable and physically reliable models for each piezometer.
In future works, we will take an interest in the challenge to forecast multiple time series with physical connections.
